\documentclass[sigconf,nonacm]{acmart}

\usepackage{algorithm}
\usepackage{algorithmic}
\usepackage{xspace}
\usepackage{multirow}

\newcommand{\pinkurl}[1]{%
  \href{#1}{\textcolor{magenta}{\nolinkurl{#1}}}%
}

\AtBeginDocument{%
}

\setcopyright{none}
\begin{document}

% ------------------------------------------------------------------
% Title
% ------------------------------------------------------------------

\title{Detail Continuation over a Trustworthy Coarse Scale for Autoregressive Super-Resolution}

% ------------------------------------------------------------------
% Authors
% ------------------------------------------------------------------

\author{Hongyi Fang}
\authornote{This work was done during an internship at Sun Yat-Sen University.}
\orcid{0009-0002-2284-5859}
\affiliation{%
  \institution{Sun Yat-Sen University}
  \city{Zhuhai}
  \country{China}
}
\affiliation{%
  \institution{Beijing Institute of Technology}
  \city{Zhuhai}
  \country{China}
}
\email{hyfang@bit.edu.cn}

\author{Jiahui Wu}
\orcid{0009-0003-6016-6301}
\affiliation{%
  \institution{Sun Yat-Sen University}
  \city{Zhuhai}
  \country{China}
}
\email{wujh287@mail2.sysu.edu.cn}

\author{Yichen Yue}
\orcid{0009-0009-7239-9464}
\affiliation{%
  \institution{Beijing Institute of Technology}
  \city{Zhuhai}
  \country{China}
}
\affiliation{%
  \institution{Sun Yat-Sen University}
  \city{Zhuhai}
  \country{China}
}
\email{yichenyue@bit.edu.cn}

\author{Benjia Zhou}
\orcid{0000-0003-4883-5552}
\affiliation{%
  \institution{Beijing Institute of Technology}
  \city{Zhuhai}
  \country{China}
}
\email{zhoubenjia@bitzh.edu.cn}

\author{Dan Zeng}
\authornote{Corresponding author.}
\orcid{0000-0002-9036-7791}
\affiliation{%
  \institution{Sun Yat-Sen University}
  \city{Zhuhai}
  \country{China}
}
\affiliation{%
  \institution{Technology Innovation Center for Collaborative Applications
    of Natural Resources Data in GBA, Ministry of Natural Resources}
  \city{Guangzhou}
  \country{China}
}
\email{zengd8@mail.sysu.edu.cn}

\renewcommand{\shortauthors}{Hongyi Fang, Jiahui Wu, Yichen Yue, Benjia Zhou \& Dan Zeng}

% ------------------------------------------------------------------
% Abstract
% ------------------------------------------------------------------

\begin{abstract}
Hallucination remains a persistent challenge in generative super-resolution (GSR), where reconstructed results may contain visually plausible yet weakly supported content, structural deviations, or unnatural textures with respect to the low-resolution (LR) input. Existing GSR methods have extensively explored the trade-off between perceptual realism and reconstruction fidelity, but the division between preserving reliable coarse-scale information and restoring more uncertain fine details is often handled implicitly within the overall restoration process. Visual autoregressive (VAR) modeling provides a natural opportunity to revisit this issue, as its coarse-to-fine next-scale prediction offers an explicit scale-wise generation interface. However, existing VAR-based SR methods still inherit the original full 1-to-$N$ autoregressive generation path, even though, for super-resolution, coarse-scale information in LR is often relatively more reliable, while long autoregressive chains may accumulate prediction errors. Motivated by these observations, we propose \textbf{K2N}, which reformulates VAR-based SR from full-path generation into a $k$-to-$N$ detail continuation process. Specifically, early coarse-scale states are established directly from LR, while only the remaining finer scales are restored autoregressively. Experimental results show that K2N remains competitive with the VARSR baseline on standard SR metrics, while exhibiting clearer advantages on hallucination-focused evaluation. These findings suggest that explicitly rethinking the generation path in a scale-wise manner can be a promising direction for improving the reliability of generative super-resolution.Our code will be released soon at:
\pinkurl{https://github.com/BRL-SYSU/K2NSR}.
\end{abstract}

% ------------------------------------------------------------------
% Keywords
% ------------------------------------------------------------------

\keywords{
  image super-resolution,
  visual autoregressive modeling,
  hallucination mitigation
}

\maketitle

% ------------------------------------------------------------------
% Overview figure
% ------------------------------------------------------------------

\begin{figure}[htbp]
  \centering
  \includegraphics[width=0.8\columnwidth]{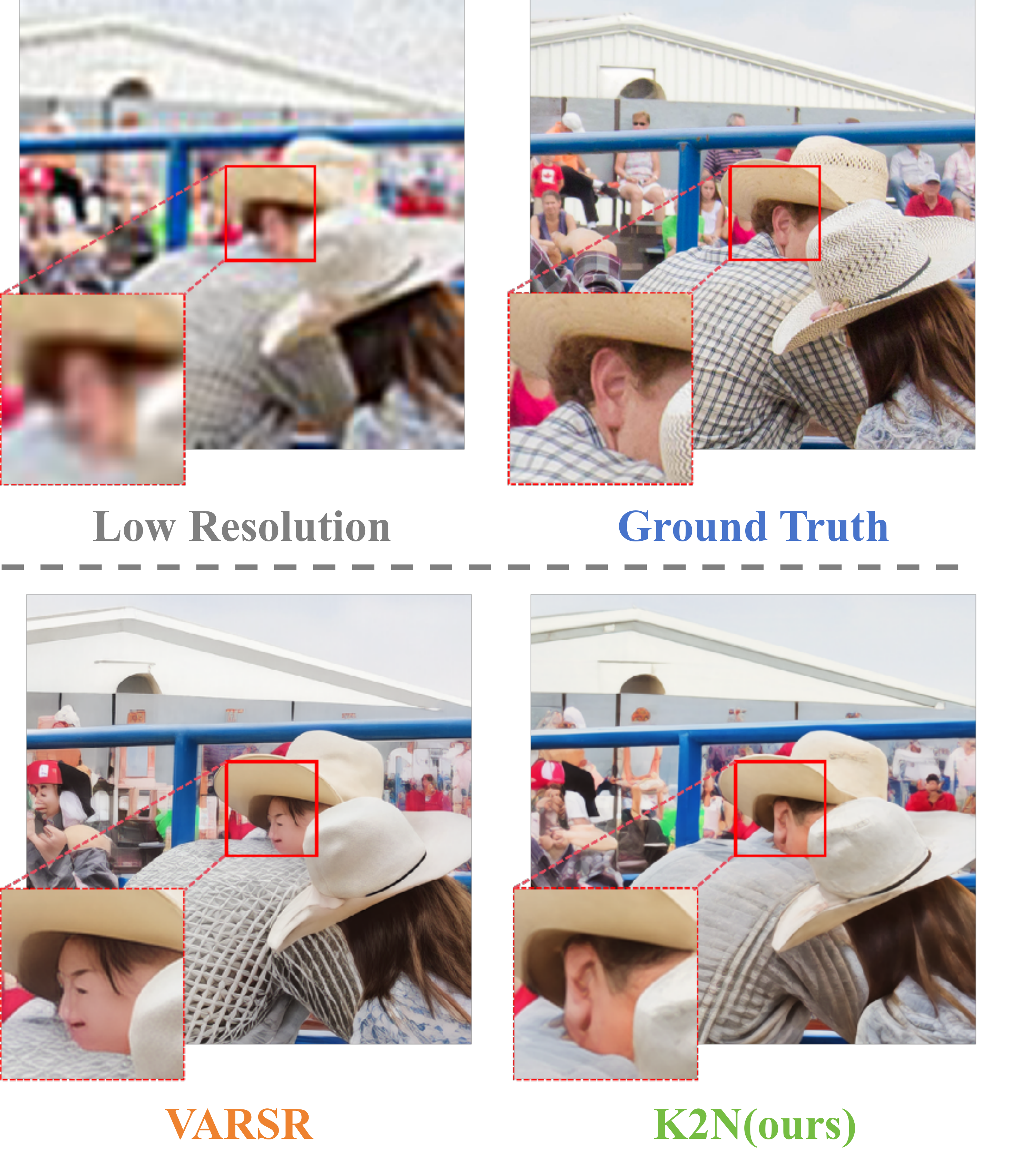}
  \caption{Visual example of hallucination mitigation.}
  \Description{%
    Side-by-side super-resolution crops compare a hallucination-prone
    reconstruction with K2N; highlighted regions show that K2N better
    preserves structures supported by the low-resolution input.
  }
  \label{fig:hallucination-mitigation}
\end{figure}

% ------------------------------------------------------------------
% Main sections
% ------------------------------------------------------------------

\section{Introduction}
Generative super-resolution (GSR) aims to restore high-resolution (HR) images with realistic and detailed content from degraded low-resolution (LR) inputs. Benefiting from recent advances in generative modeling, GSR has achieved substantial progress in visual quality and detail synthesis\cite{qu2025visualautoregressivemodelingimage,chen2024faithdiffunleashingdiffusionpriors}. Nevertheless, hallucination remains a key challenge in this setting: reconstructed results may contain details, structural deviations, or textures that appear plausible but are weakly supported by the LR observation\cite{chen2024faithdiffunleashingdiffusionpriors,ren2026hallucinationscoremitigatinghallucinations}. 

To this end, existing GSR methods have devoted considerable effort to balancing perceptual realism and reconstruction fidelity\cite{Blau_2018}. In particular, diffusion-based approaches have demonstrated strong restoration capability through iterative denoising over global latent representations, often combined with conditional guidance\cite{ho2022classifierfreediffusionguidance} and multi-scale feature modeling\cite{qu2024xpsrcrossmodalpriorsdiffusionbased}, without relying on an explicit scale-wise generation process. To further improve controllability\cite{hu2026gemtalk}, prior works have also explored hierarchical condition injection\cite{zhang2023addingconditionalcontroltexttoimage}, and semantic guidance\cite{ye2023ipadaptertextcompatibleimage,hu2026semi} to better utilize image representations at different levels. However, the core restoration process of these methods is still largely built upon a holistic denoising trajectory, where control is mainly exerted through continuous adjustment of the overall sampling process, rather than through an explicitly defined scale-wise restoration path. As a result, under the diffusion paradigm, the division of labor between coarse-scale information and fine-detail generation is usually implemented implicitly, making it less straightforward to separate restoration into two clear stages: first establishing stable coarse-scale representations from reliable LR evidence, and then recovering more uncertain fine details.

In contrast, visual autoregressive (VAR)\cite{tian2024visualautoregressivemodelingscalable} modeling enables explicit coarse-to-fine scale-wise generation, offering a natural framework for rethinking super-resolution from a scale-wise perspective. At the same time, autoregressive generation also suffers from error accumulation: each prediction step is conditioned on all previously generated scales, inaccuracies introduced at early stages can be propagated to later steps and accumulate over a long generation chain\cite{zhou2025rethinkingtrainingdynamicsscalewise}. Crucially, in super-resolution, such error accumulation is partly avoidable. Unlike unconstrained generation, the LR input still preserves reliable coarse-scale information under various degradations, with relatively small discrepancies from its HR counterpart at early scales (see Figure~2). This raises a natural question: \emph{when the LR input already provides strong support for early coarse scales, is the original full 1-to-$N$ generation path still the most suitable formulation for SR?}

Motivated by this question, we propose \textbf{K2N}, a VAR-based super-resolution framework that reformulates full-path generation as $k$-to-$N$ detail continuation. Specifically, we introduce a parallel coarse-scale reconstruction module that directly predicts the early low-scale states from the LR image, so that stable and reliable coarse-scale representations can be established before the autoregressive decoder proceeds to later fine-detail restoration. The remaining, more uncertain fine-scale components are then continued by the original VAR decoding process. In this way, K2N reduces unnecessary early autoregressive prediction while preserving the generative flexibility of VAR where it is most needed. Experimental results show that K2N performs favorably against the VARSR\cite{qu2025visualautoregressivemodelingimage} baseline on standard SR metrics, while exhibiting clearer advantages on hallucination-focused evaluation. Overall, these findings suggest that reformulating VAR-based SR as detail continuation can be a promising direction for improving reliability in generative super-resolution.

Our contributions are summarized as follows:
\begin{itemize}
    \item We point out that, although the division between coarse-scale constraints and fine-detail restoration has been widely considered in GSR, it is usually implemented implicitly in existing restoration paradigms; by contrast, the scale-wise generation mechanism of VAR enables this division to be modeled more explicitly.
    \item We propose \textbf{K2N}, a VAR-based framework that replaces the full autoregressive generation of early scales with LR-conditioned parallel coarse-scale reconstruction, reformulating SR as a $k$-to-$N$ detail continuation process.
    \item We construct a 100-image hallucination-focused diagnostic evaluation set and show that K2N achieves stronger hallucination mitigation than representative baselines while performing favorably on standard SR benchmarks.
\end{itemize}
\begin{figure}[t]
    \centering
    \includegraphics[width=\columnwidth]{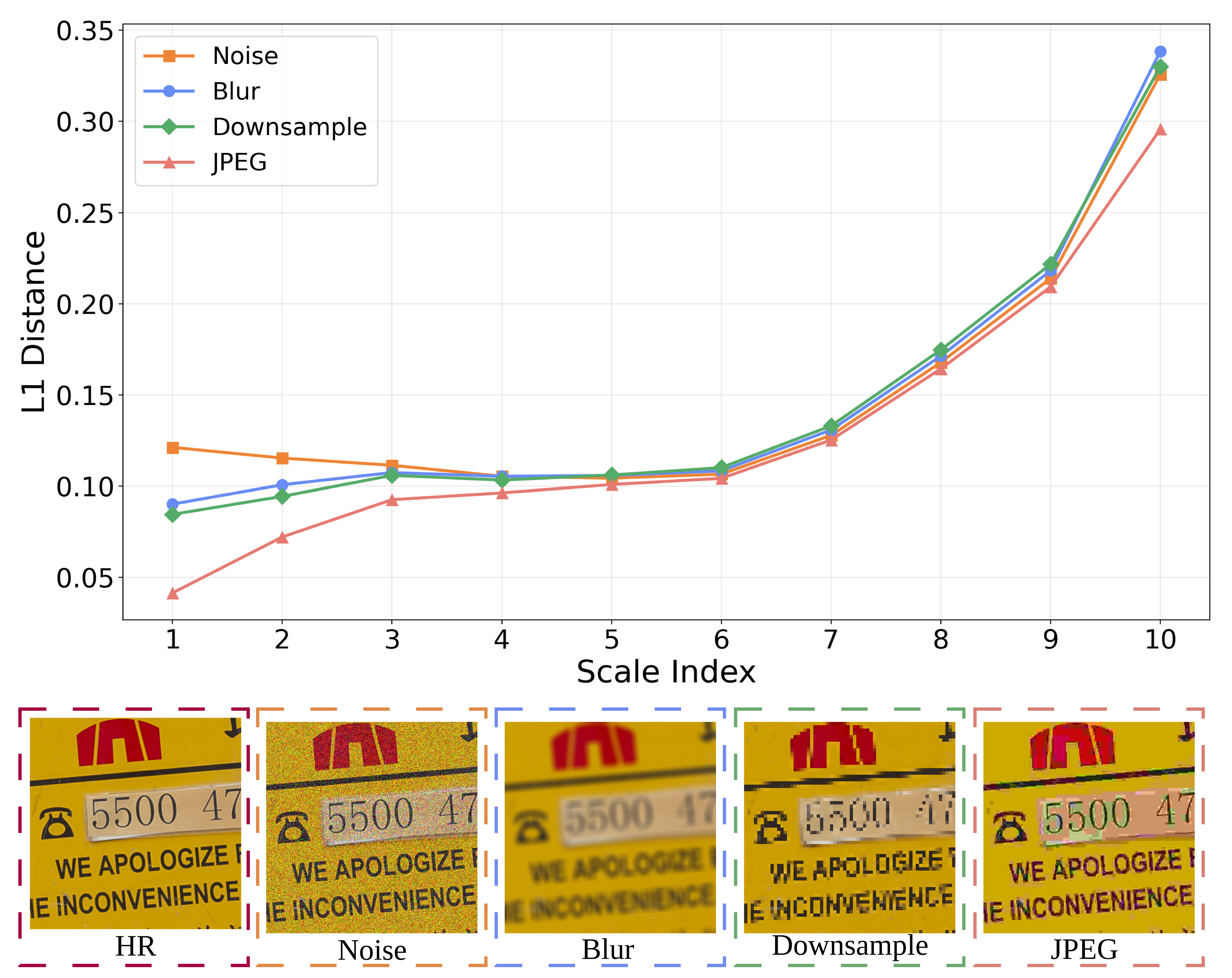}
    \caption{L1 distance between degraded and HR residual features across scales under different degradations. Early scales exhibit smaller discrepancies and are less affected by degradation, supporting our design that directly anchors coarse-scale states to LR evidence.}
    \label{fig:curve}
\end{figure}

\begin{figure*}[t]
  \centering
  \includegraphics[width=\textwidth]{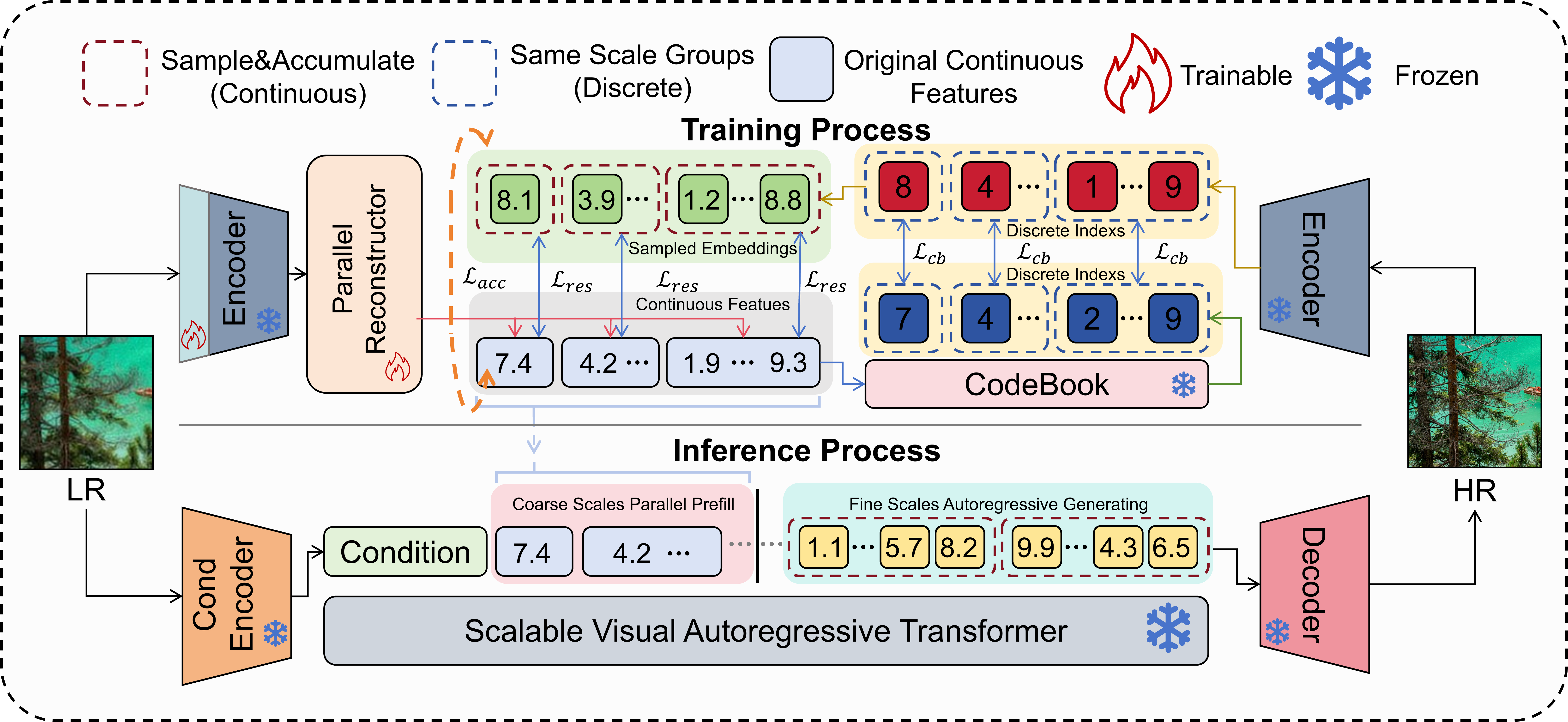} 
  \caption{
Overview of K2N during training and inference.
K2N reformulates VAR-based super-resolution from full-path autoregressive generation into a coarse-to-fine continuation process: it first predicts the coarse-scale prefix from the LR input, and then uses the frozen VAR Transformer to generate the remaining finer scales before decoding the final HR image.
}
  \label{fig:method}
\end{figure*}

\section{Related Work}

\subsection{Image Super-Resolution}

Image super-resolution has been extensively studied for decades. Early methods mainly focused on deterministic reconstruction under predefined degradations such as bicubic downsampling~\cite{dong2015imagesuperresolutionusingdeep,
liang2021swinirimagerestorationusing,
lim2017enhanceddeepresidualnetworks,
zhang2018imagesuperresolutionusingdeep}. Although these methods achieve strong distortion-oriented performance, they often fail to recover realistic textures under real-world degradations. To improve perceptual quality, later approaches introduced adversarial learning to better match natural image distributions~\cite{
wang2021realesrgantrainingrealworldblind,
zhang2019ranksrgangenerativeadversarialnetworks,
zhang2021designingpracticaldegradationmodel}, leading to substantially sharper and more visually pleasing results. More recently, diffusion-based super-resolution methods have achieved impressive perceptual quality by leveraging large-scale pretrained generative priors~\cite{wang2024exploitingdiffusionpriorrealworld, chen2024faithdiffunleashingdiffusionpriors, wu2024seesrsemanticsawarerealworldimage, lin2024diffbirblindimagerestoration}. These methods formulate super-resolution as conditional image generation and progressively refine the output through iterative denoising. At the same time, many studies have explored realism--fidelity trade-off mechanisms to better balance visual plausibility and consistency with the input~\cite{
Blau_2018, chen2024faithdiffunleashingdiffusionpriors,Zhu_2024}. However, diffusion-based super-resolution is typically realized through whole-image denoising, and the division between coarse-scale anchoring and fine-detail generation is therefore usually handled implicitly within a unified sampling trajectory.

\subsection{Visual Autoregressive Modeling}

Autoregressive modeling has recently regained attention in vision due to its success in large-scale sequence modeling~\cite{yu2022scaling,tian2024visualautoregressivemodelingscalable}. Early visual autoregressive approaches typically tokenize images into discrete patches and perform next-token prediction in a rasterized order~\cite{oord2016conditionalimagegenerationpixelcnn,
oord2016pixelrecurrentneuralnetworks}, which often struggles to preserve spatial structure in high-resolution synthesis. To address this issue, recent visual autoregressive models adopt next-scale prediction, where an image is generated in a coarse-to-fine manner over multiple scales~\cite{tian2024visualautoregressivemodelingscalable,tang2024hartefficientvisualgeneration,han2025infinityscalingbitwiseautoregressive,jiao2026flexvarflexiblevisualautoregressive}, improving both structural consistency and generation quality. This multi-scale formulation has also motivated the extension of visual autoregression from image generation to image restoration~\cite{qu2025visualautoregressivemodelingimage,wang2025navigatingimagerestorationvars}. Existing autoregressive super-resolution methods show that coarse-to-fine token prediction can provide an effective balance between perceptual quality and inference efficiency, but they still inherit the original full-path generation paradigm, requiring all scales to be generated autoregressively from the coarsest one to the finest one. In contrast, our work revisits generative super-resolution from the perspective of generation-path design. We reformulate the original full-path generation process as a $k$-to-$N$ detail continuation process, where early coarse-scale states are established directly from the input image and only the remaining finer scales are restored autoregressively.

\begin{figure*}[t]
    \centering
    \includegraphics[width=\textwidth]{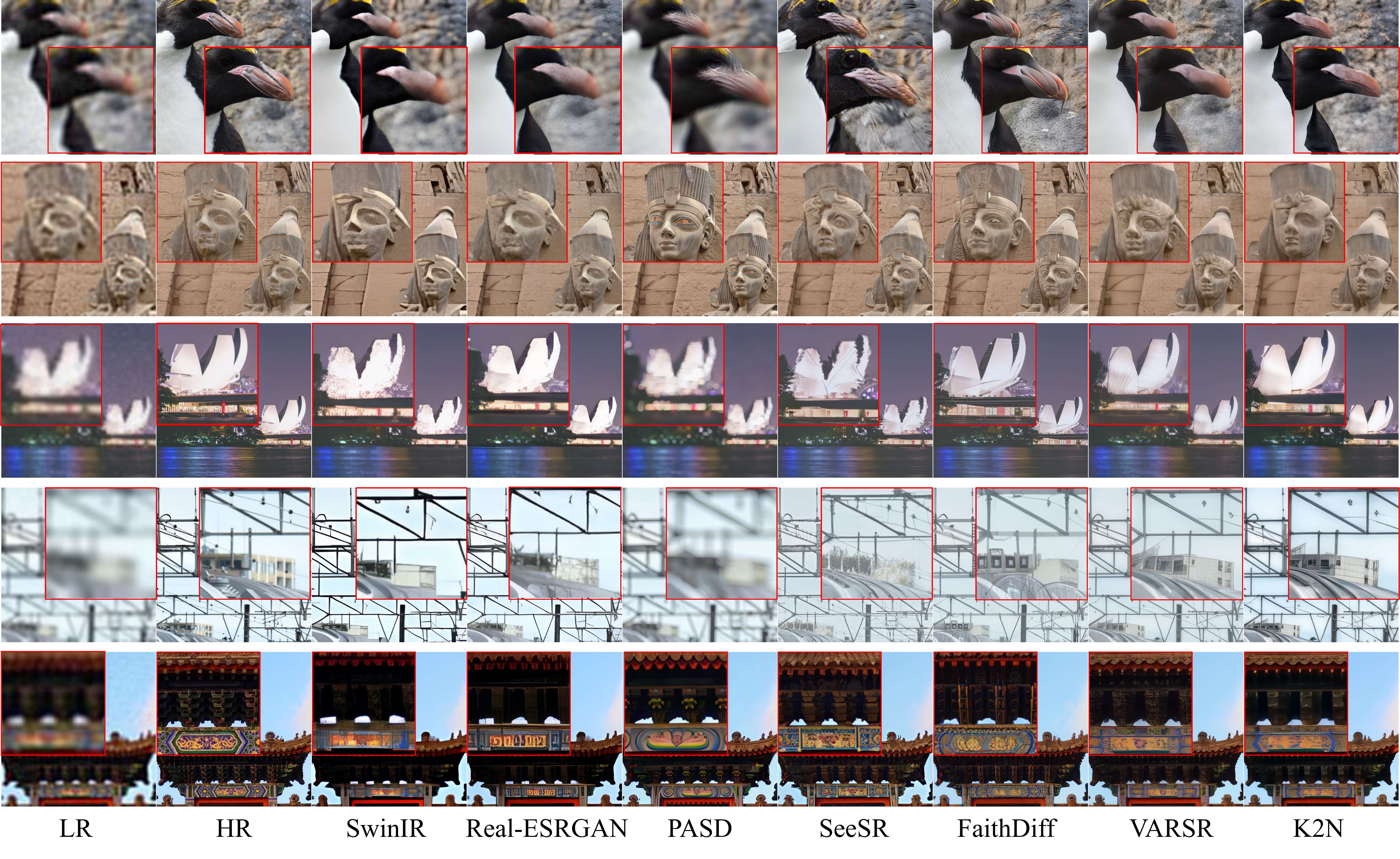}
    \caption{Qualitative comparison of different super-resolution methods. K2N restores sharper and more natural details while better preserving input-supported structures.}
    \label{fig:qualitative}
    \vspace{-5pt}
\end{figure*}

\section{Methodology}

\subsection{Preliminaries}

We follow the standard VAR tokenization pipeline and represent the HR image in a VQ latent space. Let $f$ denote the HR latent feature, $V$ the shared codebook, and $\{r_1,\dots,r_K\}$ the multi-scale token maps of the HR image, where $r_k$ denotes the discrete representation at scale $k$. Existing VAR-based SR methods typically adopt the standard next-scale autoregressive formulation:
\begin{equation}
p(r_1, r_2, \dots, r_K \mid I_{LR})
=
\prod_{k=1}^{K}
p_\theta(r_k \mid r_{<k}, I_{LR}),
\label{eq:ar}
\end{equation}
where $I_{LR}$ is the low-resolution input, and $r_{<k}$ denotes all previously generated scales before scale $k$.

Under this representation, the residual feature at scale $k$ is defined as
\begin{equation}
z_k
=
f
-
\sum_{m=1}^{k-1}
\mathrm{upsample}(\mathrm{lookup}(V, r_m)),
\label{eq:residual}
\end{equation}
where $\mathrm{lookup}(V,r_m)$ retrieves the quantized feature corresponding to $r_m$ from the shared codebook $V$, and $\mathrm{upsample}(\cdot)$ resizes it to the full latent resolution. Intuitively, $z_k$ represents the residual information that remains to be recovered after the preceding low-scale reconstructions. In the following, we use these scale-wise residual features as the main prediction target of our parallel coarse-scale reconstructor.

\begin{algorithm}[t]
\caption{\textbf{K2N}: Training and Inference}
\label{alg:k2n}
\begin{algorithmic}[1]
\REQUIRE Low-resolution image $I_{LR}$, high-resolution image $I_{HR}$ for training, prefix length $M$, total number of scales $K$
\ENSURE Super-resolved image $I_{SR}$

\STATE \textbf{Training}
\STATE $f,\{r_k\}_{k=1}^{K} \leftarrow \mathrm{VQEncode}(I_{HR})$
\FOR{$k=1$ to $M$}
    \STATE $z_k^{gt} \leftarrow f-\sum_{m=1}^{k-1}\mathrm{upsample}(\mathrm{lookup}(V,r_m))$
\ENDFOR
\STATE $\hat{z}_{1:M} \leftarrow \mathrm{PrefixRecon}(I_{LR})$
\FOR{$k=1$ to $M$}
    \STATE $\hat{s}_k \leftarrow \mathcal{A}(\hat{z}_{1:k})$
    \STATE $s_k^{gt} \leftarrow \mathcal{A}(z_{1:k}^{gt})$
    \STATE $P_k \leftarrow \mathrm{CodebookProj}(\hat{z}_k, V)$
\ENDFOR
\STATE $\mathcal{L}_{res} \leftarrow \sum_{k=1}^{M}\|\hat{z}_k-z_k^{gt}\|_1$
\STATE $\mathcal{L}_{cb} \leftarrow \sum_{k=1}^{M}\sum_{(i,j)}\mathrm{CE}(P_k(i,j,:),r_k^{gt}(i,j))$
\STATE $\mathcal{L}_{acc} \leftarrow \sum_{k=1}^{M}\|\hat{s}_k-s_k^{gt}\|_1$
\STATE $\mathcal{L}_{total} \leftarrow \mathcal{L}_{res}+\lambda\mathcal{L}_{cb}+\beta\mathcal{L}_{acc}$
\STATE Update the LR encoder and the parallel coarse-scale reconstructor using $\mathcal{L}_{total}$

\STATE \textbf{Inference}
\STATE $\hat{z}_{1:M} \leftarrow \mathrm{PrefixRecon}(I_{LR})$
\FOR{$k=1$ to $M$}
    \STATE $\hat{s}_k \leftarrow \mathcal{A}(\hat{z}_{1:k})$
\ENDFOR
\STATE $\mathcal{T} \leftarrow \mathrm{Prefill}(\{\hat{s}_k\}_{k=1}^{M})$
\FOR{$k=M+1$ to $K$}
    \STATE $r_k \leftarrow p_\theta(\cdot \mid r_{M+1:k-1}, \hat{s}_{1:M}, I_{LR})$
\ENDFOR
\STATE $I_{SR} \leftarrow \mathrm{Decode}(\{r_k\}_{k=M+1}^{K}, \{\hat{s}_k\}_{k=1}^{M})$
\RETURN $I_{SR}$
\end{algorithmic}
\end{algorithm}

\subsection{Overview}
K2N reformulates VAR-based super-resolution from full-path $1$-to-$K$ autoregressive generation into a $k$-to-$N$ detail continuation process. 
Instead of generating all scales autoregressively, we first predict the first $M$ coarse-scale prefix states in parallel from the LR input, and then use them as the prefix for subsequent VAR continuation. The pretrained VAR Transformer is thus applied only to the remaining finer scales, where generative modeling is more necessary. Figure~\ref{fig:method} illustrates the overall framework. During training, the coarse-scale prefix predictor is learned with HR-derived supervision. 
During inference, the predicted coarse prefix is pre-filled into the frozen VAR Transformer for fine-detail generation. 
The following subsections detail the continuation reformulation and prefix prediction module, with the overall procedure summarized in Algorithm~\ref{alg:k2n}.

\subsection{From Full-Path Generation to $k$-to-$N$ Detail Continuation}

The full 1-to-$N$ autoregressive generation path in Eq.~\ref{eq:ar} is inherited from generic image generation, where all scales need to be synthesized step by step from scratch. For super-resolution, however, the LR input usually already preserves relatively reliable coarse-scale structural information, making the early coarse scales less dependent on full autoregressive generation. Meanwhile, autoregressive generation also suffers from error accumulation: since each prediction step is conditioned on all previously generated scales, inaccuracies introduced at early stages can be propagated to later steps and accumulate over a longer generation chain.

Motivated by this observation, we no longer generate the first $M$ coarse scales autoregressively. Motivated by this observation, we replace the autoregressive generation of the first $M$ coarse scales with a parallel coarse-scale reconstructor that directly predicts the corresponding residual states from the LR input:
\begin{equation}
p(z_{1:M}\mid I_{LR}) \approx p_\phi(z_{1:M}\mid I_{LR}),
\label{eq:joint_prefix}
\end{equation}
where $p_\phi$ jointly predicts the first $M$ continuous residual features from LR evidence, which then serve as the starting prefix for subsequent fine-detail continuation.

Since VAR builds its generation context through scale-wise accumulated prefixes rather than raw residual features, we further construct scale-wise prefix states from the predicted residuals:
\begin{equation}
\hat{s}_k = \mathcal{A}(\hat{z}_{1:k}),
\label{eq:prefix_construct}
\end{equation}
where $\mathcal{A}(\cdot)$ denotes the scale-wise accumulation operator used to construct the transformer prefix. After obtaining the accumulated prefix states $\hat{s}_{1:M}$, the remaining finer scales are still generated by the original VAR Transformer:
\begin{equation}
p(r_{M+1},\dots,r_K \mid I_{LR}, \hat{s}_{1:M})
=
\prod_{k=M+1}^{K}
p_\theta(r_k \mid r_{<k}, \hat{s}_{1:M}, I_{LR}),
\label{eq:continuation}
\end{equation}
where the early coarse-scale context is provided by the predicted prefix states $\hat{s}_{1:M}$, while the remaining finer scales are generated autoregressively.

In this way, the original full-path autoregressive generation is reformulated as a detail continuation process starting from coarse-scale prefix states. In other words, we rewrite the original 1-to-$N$ generation mechanism in SR as a $k$-to-$N$ detail continuation process: the early coarse scales are directly established by an LR-conditioned parallel estimator, while the later, more uncertain fine scales are restored by the original autoregressive transformer. \textbf{This design shortens the effective autoregressive path and provides a more stable prefix state for subsequent fine-detail generation.}

\begin{table*}[t]
\centering
\caption{Comparison with SOTA methods on synthetic and real-world benchmarks.
\textcolor{red}{Red} and \textcolor{blue}{blue} denote the best and second-best results.}
\label{tab:main_results}
\resizebox{\textwidth}{!}{
\begin{tabular}{l l | c c c | c c c c c | c c}
\toprule
\multirow{2}{*}{Dataset} & \multirow{2}{*}{Metrics}
& \multicolumn{3}{c|}{GAN-based}
& \multicolumn{5}{c|}{Diffusion-based}
& \multicolumn{2}{c}{AR-based} \\
& & SwinIR & Real-ESRGAN & BSRGAN
& StableSR & LDM & DiffBIR & SeeSR & FaithDiff
& VARSR & \textbf{K2N} \\
\midrule

\multirow{8}{*}{DIV2K-Val}
& PSNR$\uparrow$
& 23.93 & \textcolor{blue}{24.27} & \textcolor{red}{24.58} & 23.28 & 23.32 & 20.62 & 23.68 & 23.36 & 23.99 & 23.81 \\
& SSIM$\uparrow$
& 0.6245 & \textcolor{red}{0.6338} & \textcolor{blue}{0.6269} & 0.5740 & 0.5755 & 0.5105 & 0.6042 & 0.5770 & 0.6096 & 0.6168 \\
& LPIPS$\downarrow$
& 0.4147 & 0.4080 & 0.3351 & \textcolor{red}{0.3111} & 0.3214 & 0.3967 & 0.3195 & \textcolor{blue}{0.3132} & 0.3176 & 0.3281 \\
& DISTS$\downarrow$
& 0.2143 & 0.2140 & 0.2275 & 0.2045 & \textcolor{red}{0.1961} & 0.2325 & \textcolor{blue}{0.1969} & 0.2001 & 0.2176 & 0.2143 \\
& NIQE$\downarrow$
& 4.80 & 4.78 & \textcolor{red}{4.75} & \textcolor{blue}{4.78} & 5.58 & 5.16 & 4.81 & 5.00 & 5.98 & 5.54 \\
& CLIPIQA$\uparrow$
& 0.5336 & 0.5278 & 0.5246 & 0.6753 & 0.6273 & 0.6950 & 0.6937 & 0.6523 & \textcolor{blue}{0.7337} & \textcolor{red}{0.7418} \\
& MUSIQ$\uparrow$
& 60.22 & 61.06 & 61.19 & 65.69 & 62.36 & 68.10 & 68.64 & 69.24 & \textcolor{blue}{71.49} & \textcolor{red}{71.67} \\
& MANIQA$\uparrow$
& 0.3514 & 0.3655 & 0.3532 & 0.4186 & 0.3731 & 0.4800 & 0.5034 & 0.4291 & \textcolor{blue}{0.5175} & \textcolor{red}{0.5208} \\
\midrule

\multirow{8}{*}{RealSR}
& PSNR$\uparrow$
& \textcolor{blue}{26.31} & 25.69 & \textcolor{red}{26.38} & 24.60 & 25.65 & 20.04 & 25.15 & 25.23 & 25.57 & 24.98 \\
& SSIM$\uparrow$
& \textcolor{red}{0.7731} & 0.7615 & \textcolor{blue}{0.7651} & 0.7057 & 0.7224 & 0.5393 & 0.7211 & 0.7029 & 0.7267 & 0.7283 \\
& LPIPS$\downarrow$
& 0.3574 & 0.3711 & \textcolor{red}{0.2656} & 0.3069 & 0.3033 & 0.4375 & 0.3007 & \textcolor{blue}{0.2926} & 0.3232 & 0.3155 \\
& DISTS$\downarrow$
& \textcolor{red}{0.1946} & \textcolor{blue}{0.2061} & 0.2124 & 0.2166 & 0.2208 & 0.2748 & 0.2224 & 0.2127 & 0.2358 & 0.2307 \\
& NIQE$\downarrow$
& 5.80 & 5.93 & 5.64 & 5.91 & 6.56 & 6.60 & \textcolor{red}{5.40} & \textcolor{blue}{5.45} & 6.05 & 5.72 \\
& CLIPIQA$\uparrow$
& 0.4367 & 0.4484 & 0.5114 & 0.6211 & 0.5807 & 0.6681 & 0.6697 & 0.6153 & \textcolor{blue}{0.7006} & \textcolor{red}{0.7036} \\
& MUSIQ$\uparrow$
& 58.69 & 60.37 & 63.28 & 65.30 & 59.58 & 68.42 & 69.82 & 68.67 & \textcolor{blue}{71.30} & \textcolor{red}{71.37} \\
& MANIQA$\uparrow$
& 0.3514 & 0.3758 & 0.3758 & 0.4213 & 0.3686 & 0.5124 & 0.5437 & 0.4657 & \textcolor{blue}{0.5480} & \textcolor{red}{0.5481} \\
\midrule

\multirow{8}{*}{DRealSR}
& PSNR$\uparrow$
& 28.50 & \textcolor{blue}{28.61} & \textcolor{red}{28.70} & 27.97 & 28.07 & 22.50 & 28.07 & 27.26 & 28.15 & 27.18 \\
& SSIM$\uparrow$
& \textcolor{blue}{0.8041} & \textcolor{red}{0.8050} & 0.8028 & 0.7500 & 0.7526 & 0.5494 & 0.7684 & 0.7099 & 0.7645 & 0.7652 \\
& LPIPS$\downarrow$
& 0.3692 & 0.3766 & \textcolor{red}{0.2858} & 0.3317 & 0.3231 & 0.5209 & \textcolor{blue}{0.3174} & 0.3523 & 0.3542 & 0.3500 \\
& DISTS$\downarrow$
& \textcolor{red}{0.2040} & \textcolor{blue}{0.2089} & 0.2144 & 0.2310 & 0.2210 & 0.3077 & 0.2315 & 0.2388 & 0.2535 & 0.2480 \\
& NIQE$\downarrow$
& 6.79 & 6.86 & 6.54 & 6.56 & 6.92 & 7.70 & \textcolor{red}{6.40} & \textcolor{blue}{6.43} & 6.94 & 6.55 \\
& CLIPIQA$\uparrow$
& 0.4444 & 0.4515 & 0.5091 & 0.6226 & 0.5681 & 0.6560 & 0.6909 & 0.6300 & \textcolor{blue}{0.7224} & \textcolor{red}{0.7287} \\
& MUSIQ$\uparrow$
& 52.73 & 54.28 & 57.16 & 58.98 & 53.79 & 60.70 & 65.08 & 66.35 & \textcolor{blue}{68.05} & \textcolor{red}{68.99} \\
& MANIQA$\uparrow$
& 0.3244 & 0.3388 & 0.3403 & 0.3897 & 0.3353 & 0.4533 & 0.5129 & 0.4501 & \textcolor{blue}{0.5356} & \textcolor{red}{0.5489} \\
\bottomrule
\end{tabular}}
\end{table*}

\subsection{Coarse-Scale Prefix Prediction and Supervision}

A direct choice for the first $M$ scales is to predict discrete tokens. Instead, we predict \textbf{continuous residual features}. Compared with quantized discrete outputs, continuous features preserve richer structural information and are therefore more suitable as coarse-scale prefix states.

Based on this design, we impose three complementary supervisions on the outputs of the parallel coarse-scale reconstructor, respectively at the residual level, the codebook level, and the accumulated prefix-state level.

First, we apply scale-wise $L_1$ reconstruction loss to the predicted continuous residual features:
\begin{equation}
\mathcal{L}_{res}
=
\sum_{k=1}^{M}
\left\|
\hat{z}_k - z_k^{gt}
\right\|_1,
\label{eq:lres}
\end{equation}
where $\hat{z}_k$ and $z_k^{gt}$ denote the predicted and ground-truth residual features at scale $k$, respectively. This term directly constrains each prediction in the continuous feature space.

Second, to maintain compatibility with the pretrained VAR Transformer, we introduce codebook-aligned supervision. Specifically, we project each predicted feature $\hat{z}_k$ into the shared codebook space and construct a token distribution $P_k$ according to its similarity to the codebook entries. Based on the ground-truth token index $r_k^{gt}$, we define the codebook-aligned loss as
\begin{equation}
\mathcal{L}_{cb}
=
\sum_{k=1}^{M}
\sum_{(i,j)}
\mathrm{CE}\big(P_k(i,j,:), r_k^{gt}(i,j)\big).
\label{eq:lcb}
\end{equation}
While continuous residual features remain the primary prediction target, this term aligns them with the discrete representation space used by VAR for more reliable continuation generation.

Furthermore, since the predicted residual features are ultimately accumulated into transformer prefix states, we introduce an additional accumulated-state consistency supervision to regularize cross-scale coherence. Based on Eq.~\ref{eq:prefix_construct}, the ground-truth accumulated prefix state is defined as
\begin{equation}
s_k^{gt} = \mathcal{A}(z_{1:k}^{gt}),
\label{eq:acc_state}
\end{equation}
where $\mathcal{A}(\cdot)$ denotes the same accumulation operator used to construct the transformer prefix, which progressively accumulates the prefix features up to the current scale before feeding them into the continuation transformer. We then define
\begin{equation}
\mathcal{L}_{acc}
=
\sum_{k=1}^{M}
\left\|
\hat{s}_k - s_k^{gt}
\right\|_1.
\label{eq:lacc}
\end{equation}
Different from the residual-level supervision in Eq.~\ref{eq:lres}, this term directly constrains the accumulated prefix representations actually used by the continuation transformer, thereby improving consistency across scales.

The overall objective is
\begin{equation}
\mathcal{L}_{total}
=
\mathcal{L}_{res}
+
\lambda \mathcal{L}_{cb}
+
\beta \mathcal{L}_{acc},
\label{eq:total}
\end{equation}
where $\lambda$ and $\beta$ balance the codebook-aligned supervision and the accumulated-state consistency supervision, respectively.

In summary, our method uses continuous residual features as the primary prediction target, aligns them with the pretrained VAR codebook space through auxiliary discrete supervision, and further regularizes the accumulated prefix states to improve cross-scale consistency. The resulting prefix states are then pre-filled into the subsequent VAR Transformer to continue the generation of the remaining fine scales.

\section{Experiments}

\subsection{Experimental Setup}

\paragraph{Datasets.}
We train K2N on 1.6M cropped HR patches of resolution $512\times512$, sampled from a collection of 28K images from DF2K\cite{Agustsson_2017_CVPR_Workshops}, OST\cite{wang2018sftgan}, and Unsplash\_Lite. Following the blind SR setting, LR-HR training pairs are synthesized using the degradation pipeline of Real-ESRGAN\cite{wang2021realesrgantrainingrealworldblind}. Unless otherwise specified. For evaluation, we adopt both synthetic and real-world benchmarks, including DIV2K-Val\cite{Agustsson_2017_CVPR_Workshops}, RealSR\cite{Cai_2019_ICCV}, and DRealSR\cite{wei2020componentdivideandconquerrealworldimage}, to assess the proposed method under controlled degradations and realistic image corruptions. In addition, we construct a hallucination-focused evaluation set for diagnosing whether the restored result remains faithful to LR-supported content. This benchmark consists of LR-HR pairs in which the LR input still preserves recognizable semantic or structural cues. It is built through a two-stage process, including a broader automatically pre-screened candidate pool and a manually verified high-confidence subset. We report multimodal-LLM-based scoring\cite{ren2026hallucinationscoremitigatinghallucinations} in both sets and further conduct a human pairwise comparison in the manually verified subset. Full construction details are provided in the supplementary material.

\paragraph{Evaluation Metrics.}
We report both standard restoration metrics and perceptual quality metrics. For reference-based evaluation, we use PSNR, SSIM\cite{1284395}, LPIPS\cite{zhang2018unreasonableeffectivenessdeepfeatures}, and DISTS\cite{Ding_2020}. For no-reference perceptual evaluation, we adopt NIQE\cite{mittal2012making}, CLIPIQA\cite{wang2022exploringclipassessinglook}, MUSIQ\cite{ke2021musiqmultiscaleimagequality} and MANIQA\cite{yang2022maniqamultidimensionattentionnetwork}. In addition, hallucination mitigation is evaluated on the hallucination-focused benchmark using multimodal-LLM-based scoring and human preference study. Together, these metrics cover distortion fidelity, perceptual similarity, perceptual quality, and faithfulness to LR evidence.

\paragraph{Implementation Details.}
We build on a pretrained VARSR\cite{qu2025visualautoregressivemodelingimage} and keep the original autoregressive generation model unchanged. Our coarse-scale reconstruction module predicts the first $M$ residual features ${z_k}{k=1}^{M}$ from the LR input in parallel, and the predicted residuals are accumulated into prefix states. Conditioned on these prefix states, the original autoregressive model then continues the subsequent scale-wise generation to restore the remaining finer details. Unless otherwise noted, we set $M=3$, so that the first three residual scales are predicted in parallel and the remaining finer scales are generated autoregressively. We use AdamW\cite{loshchilov2019decoupledweightdecayregularization} for 20 epochs with learning rate $1\times10^{-4}$, weight decay 0, gradient clipping 1.0, and per-GPU batch size 80 on 2 H200 GPUs. The coefficients of $\mathcal{L}{res}$, $\mathcal{L}{cb}$, and $\mathcal{L}{acc}$ are set to 1.0, 0.001 and 0.2, respectively.

\paragraph{Compared Methods.}
We compare K2N with nine representative SR methods, including SwinIR\cite{liang2021swinirimagerestorationusing}, Real-ESRGAN\cite{wang2021realesrgantrainingrealworldblind}, BSRGAN\cite{zhang2021designingpracticaldegradationmodel}, StableSR\cite{wang2024exploitingdiffusionpriorrealworld}, LDM\cite{rombach2022highresolutionimagesynthesislatent}, DiffBIR\cite{lin2024diffbirblindimagerestoration}, SeeSR\cite{wu2024seesrsemanticsawarerealworldimage}, FaithDiff\cite{chen2024faithdiffunleashingdiffusionpriors} and VARSR\cite{qu2025visualautoregressivemodelingimage}. These baselines cover feed-forward restoration, GAN-based generation, diffusion-based restoration, and autoregressive super-resolution, providing a broad view of how K2N performs against alternative restoration and generation paradigms. In particular, FaithDiff is included as a diffusion-based fidelity-oriented baseline to complement perceptual generative methods in the comparison.

\subsection{Comparison with State-of-the-Art Methods}

\paragraph{Quantitative Comparisons.}
As shown in Table~\ref{tab:main_results}, On both synthetic and real-world benchmarks, K2N delivers consistent improvements in perceptual quality while maintaining competitive restoration performance. In particular, K2N achieves the best CLIPIQA, MUSIQ, and MANIQA scores on all three datasets. Some GAN-based methods obtain stronger full-reference scores on individual datasets, often due to more conservative reconstructions and relatively limited generative priors, at the cost of perceptual richness. In contrast, among methods with stronger generative priors, including diffusion-based and autoregressive approaches, K2N remains competitive on DISTS, LPIPS, and especially SSIM, indicating good fidelity and structural consistency. Compared with the base model VARSR, K2N consistently improves no-reference perceptual metrics across all three benchmarks while remaining broadly comparable on standard restoration metrics. Overall, K2N improves perceptual quality without a clear loss of reconstruction fidelity.

\paragraph{Qualitative Comparisons.}
Figure.~\ref{fig:qualitative} presents visual comparisons of five representative examples. In the first penguin example, K2N is the only method that cautiously restores the beak structure while still maintaining a well-reconstructed background. In the second sculpture example, diffusion-based methods and VARSR produce exaggerated facial structures, whereas GAN-based methods are overly conservative and leave the face blurry. In the Sydney Opera House example, K2N recovers the sharp structural layout without introducing redundant textures, and its color reproduction is also the most consistent with the HR target. In the train example, diffusion-based methods and VARSR produce relatively blurry results, suggesting weaker modeling of coarse-scale structure, while GAN-based methods generate a brighter overall sky, but fail to recover finer structures such as the train roof as reliably as K2N. In the signboard example, K2N accurately restores the coarse-scale diamond-shaped boundary and preserves better structural consistency. These observations are consistent with the quantitative results. Compared with GAN-based methods, K2N avoids overly conservative reconstruction; compared with diffusion-based methods and VARSR, it better preserves LR-supported coarse structures while reducing unsupported or exaggerated details. This behavior aligns well with our design motivation of anchoring reliable coarse-scale information before continuing finer-scale generation.

\subsection{Hallucination Evaluation}

\paragraph{Multimodal Large-Model Evaluation.}
Following the hallucination evaluation protocol of Hallucination Scores~\cite{ren2026hallucinationscoremitigatinghallucinations}, we use a multimodal large model as a judge on our proposed 100-image hallucination-focused set. For each example, the model receives the GT image, the LR input, and the SR result, and is asked to assign a hallucination score based on whether the restored content introduces semantically inconsistent or perceptually disturbing details that are not supported by the input. As shown in Table~\ref{tab:mllm_eval}, K2N achieves the best score among the compared methods, indicating improved hallucination mitigation under this diagnostic protocol. Full prompting details and scoring instructions are provided in the supplementary material.

\begin{table}[t]
    \centering
    \caption{MLLM-based hallucination evaluation on the 100-image benchmark. Higher is better.}
    \label{tab:mllm_eval}
    \begin{tabular}{lcccc}
        \toprule
        Metric & VARSR & SeeSR & FaithDiff & K2N \\
        \midrule
        Hallucination Score $\uparrow$ & 3.05 & 2.93 & 3.19 & \textbf{3.31} \\
        \bottomrule
    \end{tabular}
\end{table}

\begin{figure*}[t]
    \centering
    \includegraphics[width=\textwidth]{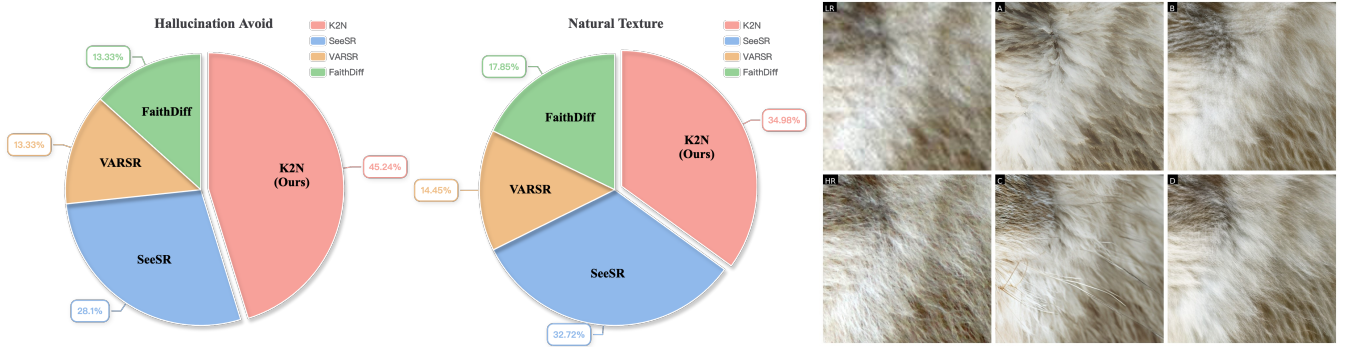}
    \caption{User study results and blind pairwise comparison examples on the hallucination-focused benchmark.}
    \label{fig:user_study}
\end{figure*}

\paragraph{User Study.}
We further conduct a user study on our hallucination-focused evaluation set. We build 45 comparison groups, randomly display 10 groups in each questionnaire, and collect 40 valid responses in total. Participants are asked to judge SR results in terms of texture naturalness and faithfulness to the LR input. As shown in Figure.~\ref{fig:user_study}, K2N is consistently preferred over the original autoregressive baseline and other representative generative SR methods. These results further support that the proposed scale-wise design improves both perceptual naturalness and faithfulness to LR evidence from a human perception perspective. Full questionnaire details are provided in the supplementary material.

% \begin{table}[t]
%     \centering
%     \caption{Human pairwise comparison on the hallucination-focused evaluation set. Higher is better.}
%     \label{tab:user_study}
%     \begin{tabular}{lcccc}
%         \toprule
%         Metric & VARSR & SeeSR & Hero-SR & K2N \\
%         \midrule
%         Hallucination Avoidance $\uparrow$ & -- & -- & -- & -- \\
%         Visual Naturalness $\uparrow$     & -- & -- & -- & -- \\
%         \bottomrule
%     \end{tabular}
% \end{table}

\subsection{Ablation Study}

\paragraph{Continuous vs. Discrete Prefix Prediction.}
We compare different forms of low-scale prefix prediction in Table~\ref{tab:feature_ablation}. Directly predicting discrete tokens for early scales leads to inferior performance, indicating that low-scale guidance should preserve richer structural information. Predicting continuous residual features yields better reconstruction and perceptual quality, and further introducing codebook-aligned supervision achieves the best overall results. This supports our design choice of using continuous features as the primary prediction target while retaining compatibility with the pretrained VAR representation space.

\begin{table}[t]
    \centering
    \caption{Ablation of different prefix prediction forms on DIV2K-Val.}
    \label{tab:feature_ablation}
    \resizebox{\columnwidth}{!}{
    \begin{tabular}{lcccc}
        \toprule
        Prefix Type & PSNR $\uparrow$ & SSIM $\uparrow$ & MANIQA $\uparrow$ & CLIPIQA $\uparrow$ \\
        \midrule
        Discrete tokens & 23.59 & \textbf{0.5924} & 0.4616 & 0.6713 \\
        Continuous features & \textbf{23.65} & 0.5920 & \textbf{0.4843} & \textbf{0.7203} \\
        \bottomrule
    \end{tabular}
    }
\end{table}

% \begin{table*}[t]
%     \centering
%     \scriptsize
%     \setlength{\tabcolsep}{4pt}
%     \caption{Ablation study of different supervision terms on DIV2K-VAL.}
%     \label{tab:loss_ablation}
%     \resizebox{0.9\textwidth}{!}{
%     \begin{tabular}{ccc|cccccccc}
%         \toprule
%         $\mathcal{L}_{res}$ & $\mathcal{L}_{cb}$ & $\mathcal{L}_{acc}$
%         & PSNR $\uparrow$
%         & SSIM $\uparrow$
%         & LPIPS $\downarrow$
%         & DISTS $\downarrow$
%         & NIQE $\downarrow$
%         & CLIP-IQA $\uparrow$
%         & MUSIQ $\uparrow$
%         & MANIQA $\uparrow$ \\
%         \midrule
%         $\checkmark$ & $\times$ & $\times$
%         & 23.65 & 0.5920 & 0.3276 & 0.2265 & 6.085 & 0.7203 & 70.14 & 0.4843 \\
%         $\checkmark$ & $\checkmark$ & $\times$
%         & 23.76 & 0.5956 & \textbf{0.3152} & 0.2206 & 5.912 & 0.7353 & 70.74 & 0.5105 \\
%         $\checkmark$ & $\times$ & $\checkmark$
%         & 23.70 & 0.5975 & 0.3254 & 0.2232 & 6.045 & 0.7410 & 71.20 & 0.5207 \\
%         $\checkmark$ & $\checkmark$ & $\checkmark$
%         & \textbf{23.81} & \textbf{0.6168} & 0.3281 & \textbf{0.2143} & \textbf{5.540} & \textbf{0.7418} & \textbf{71.67} & \textbf{0.5208} \\
%         \bottomrule
%     \end{tabular}
%     }
% \end{table*}

\begin{table}[t]
    \centering
    \caption{Ablation of supervision terms on DIV2K-Val.}
    \label{tab:loss_ablation}
    \resizebox{\columnwidth}{!}{
    \begin{tabular}{ccc|cccc}
        \toprule
        $\mathcal{L}_{res}$ & $\mathcal{L}_{cb}$ & $\mathcal{L}_{acc}$
        & PSNR $\uparrow$
        & SSIM $\uparrow$
        & MANIQA $\uparrow$
        & CLIP-IQA $\uparrow$ \\
        \midrule
        $\checkmark$ & $\times$ & $\times$
        & 23.65 & 0.5920 & 0.4843 & 0.7203 \\
        $\checkmark$ & $\checkmark$ & $\times$
        & 23.76 & 0.5956 & 0.5105 & 0.7353 \\
        $\checkmark$ & $\times$ & $\checkmark$
        & 23.70 & 0.5975 & 0.5207 & 0.7410 \\
        $\checkmark$ & $\checkmark$ & $\checkmark$
        & \textbf{23.81}
        & \textbf{0.6168}
        & \textbf{0.5208}
        & \textbf{0.7418} \\
        \bottomrule
    \end{tabular}
    }
\end{table}

\paragraph{Effect of Different Supervision Terms.}
We further analyze the role of the three supervision terms in the training objective. Specifically, we consider four settings: using only $\mathcal{L}_{res}$, adding $\mathcal{L}_{cb}$ on top of $\mathcal{L}_{res}$, adding $\mathcal{L}_{acc}$ on top of $\mathcal{L}_{res}$, and using the full objective $\mathcal{L}_{res} + \lambda \mathcal{L}_{cb} + \beta \mathcal{L}_{acc}$. As shown in Table~\ref{tab:loss_ablation}, using only residual supervision already provides a reasonable coarse-scale estimate, but the results remain suboptimal. Adding codebook-aligned supervision improves compatibility between the predicted continuous features and the discrete representation space expected by the pretrained decoder, while accumulated-state supervision directly regularizes the constructed prefix states and improves cross-scale consistency. Combining all three losses yields the best overall balance between reconstruction quality, perceptual quality, and hallucination mitigation, validating the necessity of the full objective.

% \subsection{Analysis of Hallucination Mitigation}

% \paragraph{Visual Evidence.}
% Figure~\ref{fig:hallucination_cases} shows representative hallucination-prone cases from our diagnostic evaluation set. Existing generative SR methods may produce visually sharp but unsupported structures, especially in repeated patterns, text regions, and thin boundaries. In contrast, K2N preserves more faithful coarse geometry and reduces the introduction of excessive fake details at early stages, leading to more reliable final reconstruction.

\begin{table}[t]
\centering
\caption{Comparison of model complexity and inference cost.}
\label{tab:complexity_comparison}
\resizebox{0.8\columnwidth}{!}{
\begin{tabular}{l|c|c|c}
\toprule
Method & Params & Steps & Time (s/img) \\
\midrule
StableSR     & 1.41B & 200 & 18.91 \\
DiffBIR      & 1.72B & 50 & 5.79 \\
SeeSR        & 2.28B & 50 & 7.16 \\
VARSR        & 1.10B & 10 & 0.57 \\
\midrule
\textbf{K2N} & 1.21B & 7  & \textbf{0.51} \\
\bottomrule
\end{tabular}
}
\end{table}

\subsection{Model Complexity and Runtime Analysis}

Table~\ref{tab:complexity_comparison} compares model complexity and inference cost. 
Autoregressive methods are substantially more efficient than diffusion-based methods due to their much shorter generation paths. 
K2N maintains this efficiency advantage, requiring only 7 generation steps and achieving the lowest runtime among the compared methods, with only a modest increase in model size. 
This shows that replacing early autoregressive decoding with parallel coarse-scale prefix prediction improves efficiency in practice.

\begin{figure}[t]
    \centering
    \includegraphics[width=1\columnwidth]{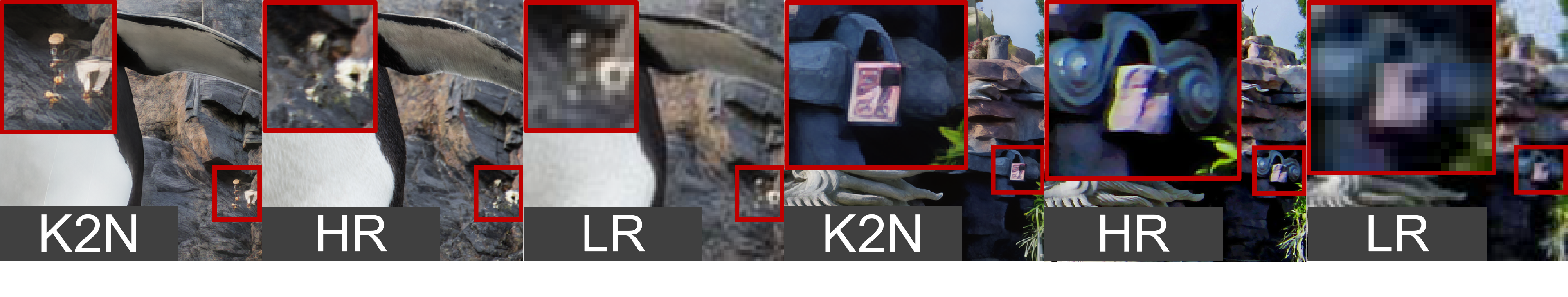}
    \caption{Typical failure cases. While K2N reduces hallucinations by anchoring coarse structures to LR evidence, errors may still occur in highly ambiguous fine-detail regions.}
    \label{fig:failure}
\end{figure}

% \paragraph{Discussion.}
% The above results support our main claim: in generative super-resolution, the realism--fidelity trade-off should be regulated in a scale-wise manner rather than uniformly across the entire generation process. By trusting coarse structures from the LR input and reserving stronger generative priors for fine-detail continuation, K2N reduces hallucinations while maintaining competitive reconstruction quality on standard benchmarks.

% ------------------------------------------------------------------
% Limitations
% ------------------------------------------------------------------

\section{Limitations and Failure Cases}

Figure~\ref{fig:failure} shows that K2N may still fail in ambiguous
fine-detail regions, incorrectly restoring the cliff flowers or introducing
unsupported facade patterns. This suggests that anchoring only the earliest
coarse scales cannot guarantee faithful detail recovery, since reliable LR
evidence may also persist at intermediate scales currently modeled
autoregressively. Extending K2N with scale-adaptive evidence preservation is
therefore a promising direction.

% ------------------------------------------------------------------
% Conclusion
% ------------------------------------------------------------------

\section{Conclusion}

We present K2N, which replaces early autoregressive SR steps with
LR-conditioned coarse-prefix prediction before autoregressive detail
continuation. By separating structural anchoring from detail generation,
K2N improves perceptual quality and reduces hallucinations while maintaining
competitive fidelity and efficiency. These results highlight scale-wise path
design as a promising approach to controllable and faithful generative
restoration.

% ------------------------------------------------------------------
% Acknowledgments
% ------------------------------------------------------------------

\section{Acknowledgments}

This work was supported by the National Natural Science Foundation of China
(Grant No.~62206123) and the Key Science and Technology Program of Lhasa
Municipality (Grant No.~LSKJ202612).

% ------------------------------------------------------------------
% References
% ------------------------------------------------------------------

\bibliographystyle{ACM-Reference-Format}
\bibliography{main}

@ARTICLE{dong2015imagesuperresolutionusingdeep,
  author={Dong, Chao and Loy, Chen Change and He, Kaiming and Tang, Xiaoou},
  journal={IEEE Transactions on Pattern Analysis and Machine Intelligence}, 
  title={Image Super-Resolution Using Deep Convolutional Networks}, 
  year={2016},
  volume={38},
  number={2},
  pages={295-307},
  doi={10.1109/TPAMI.2015.2439281}}

@InProceedings{liang2021swinirimagerestorationusing,
    author    = {Liang, Jingyun and Cao, Jiezhang and Sun, Guolei and Zhang, Kai and Van Gool, Luc and Timofte, Radu},
    title     = {SwinIR: Image Restoration Using Swin Transformer},
    booktitle = {Proceedings of the IEEE/CVF International Conference on Computer Vision (ICCV) Workshops},
    month     = {October},
    year      = {2021},
    pages     = {1833-1844}
}

@InProceedings{lim2017enhanceddeepresidualnetworks,
author = {Lim, Bee and Son, Sanghyun and Kim, Heewon and Nah, Seungjun and Mu Lee, Kyoung},
title = {Enhanced Deep Residual Networks for Single Image Super-Resolution},
booktitle = {Proceedings of the IEEE Conference on Computer Vision and Pattern Recognition (CVPR) Workshops},
month = {July},
year = {2017}
}

@InProceedings{zhang2018imagesuperresolutionusingdeep,
author = {Zhang, Yulun and Li, Kunpeng and Li, Kai and Wang, Lichen and Zhong, Bineng and Fu, Yun},
title = {Image Super-Resolution Using Very Deep Residual Channel Attention Networks},
booktitle = {Proceedings of the European Conference on Computer Vision (ECCV)},
month = {September},
year = {2018}
}

@InProceedings{zhang2019ranksrgangenerativeadversarialnetworks,
author = {Zhang, Wenlong and Liu, Yihao and Dong, Chao and Qiao, Yu},
title = {RankSRGAN: Generative Adversarial Networks With Ranker for Image Super-Resolution},
booktitle = {Proceedings of the IEEE/CVF International Conference on Computer Vision (ICCV)},
month = {October},
year = {2019}
}

@InProceedings{zhang2021designingpracticaldegradationmodel,
    author    = {Zhang, Kai and Liang, Jingyun and Van Gool, Luc and Timofte, Radu},
    title     = {Designing a Practical Degradation Model for Deep Blind Image Super-Resolution},
    booktitle = {Proceedings of the IEEE/CVF International Conference on Computer Vision (ICCV)},
    month     = {October},
    year      = {2021},
    pages     = {4791-4800}
}

@misc{wang2024exploitingdiffusionpriorrealworld,
      title={Exploiting Diffusion Prior for Real-World Image Super-Resolution}, 
      author={Jianyi Wang and Zongsheng Yue and Shangchen Zhou and Kelvin C. K. Chan and Chen Change Loy},
      year={2024},
      eprint={2305.07015},
      archivePrefix={arXiv},
      primaryClass={cs.CV},
      url={https://arxiv.org/abs/2305.07015}, 
}

@InProceedings{chen2024faithdiffunleashingdiffusionpriors,
    author    = {Chen, Junyang and Pan, Jinshan and Dong, Jiangxin},
    title     = {FaithDiff: Unleashing Diffusion Priors for Faithful Image Super-resolution},
    booktitle = {Proceedings of the IEEE/CVF Conference on Computer Vision and Pattern Recognition (CVPR)},
    month     = {June},
    year      = {2025},
    pages     = {28188-28197}
}

@InProceedings{wu2024seesrsemanticsawarerealworldimage,
    author    = {Wu, Rongyuan and Yang, Tao and Sun, Lingchen and Zhang, Zhengqiang and Li, Shuai and Zhang, Lei},
    title     = {SeeSR: Towards Semantics-Aware Real-World Image Super-Resolution},
    booktitle = {Proceedings of the IEEE/CVF Conference on Computer Vision and Pattern Recognition (CVPR)},
    month     = {June},
    year      = {2024},
    pages     = {25456-25467}
}

@InProceedings{wang2021realesrgantrainingrealworldblind,
    author    = {Wang, Xintao and Xie, Liangbin and Dong, Chao and Shan, Ying},
    title     = {Real-ESRGAN: Training Real-World Blind Super-Resolution With Pure Synthetic Data},
    booktitle = {Proceedings of the IEEE/CVF International Conference on Computer Vision (ICCV) Workshops},
    month     = {October},
    year      = {2021},
    pages     = {1905-1914}
}

@InProceedings{lin2024diffbirblindimagerestoration,
author="Lin, Xinqi
and He, Jingwen
and Chen, Ziyan
and Lyu, Zhaoyang
and Dai, Bo
and Yu, Fanghua
and Qiao, Yu
and Ouyang, Wanli
and Dong, Chao",
editor="Leonardis, Ale{\v{s}}
and Ricci, Elisa
and Roth, Stefan
and Russakovsky, Olga
and Sattler, Torsten
and Varol, G{\"u}l",
title="DiffBIR: Toward Blind Image Restoration with Generative Diffusion Prior",
booktitle="Computer Vision -- ECCV 2024",
year="2025",
publisher="Springer Nature Switzerland",
address="Cham",
pages="430--448",
isbn="978-3-031-73202-7"
}

@InProceedings{Blau_2018,
author = {Blau, Yochai and Michaeli, Tomer},
title = {The Perception-Distortion Tradeoff},
booktitle = {Proceedings of the IEEE Conference on Computer Vision and Pattern Recognition (CVPR)},
month = {June},
year = {2018}
}

@inproceedings{Zhu_2024,
author = {Zhu, Qiwen and Wang, Yanjie and Cai, Shilv and Chen, Liqun and Zhou, Jiahuan and Yan, Luxin and Zhong, Sheng and Zou, Xu},
title = {Perceptual-Distortion Balanced Image Super-Resolution is a Multi-Objective Optimization Problem},
year = {2024},
isbn = {9798400706868},
publisher = {Association for Computing Machinery},
address = {New York, NY, USA},
url = {https://doi.org/10.1145/3664647.3681512},
doi = {10.1145/3664647.3681512},
booktitle = {Proceedings of the 32nd ACM International Conference on Multimedia},
pages = {7483–7492},
numpages = {10},
location = {Melbourne VIC, Australia},
series = {MM '24}
}

@InProceedings{oord2016pixelrecurrentneuralnetworks,
  title = 	 {Pixel Recurrent Neural Networks},
  author = 	 {van den Oord, Aäron and Kalchbrenner, Nal and Kavukcuoglu, Koray},
  booktitle = 	 {Proceedings of The 33rd International Conference on Machine Learning},
  pages = 	 {1747--1756},
  year = 	 {2016},
  editor = 	 {Balcan, Maria Florina and Weinberger, Kilian Q.},
  volume = 	 {48},
  series = 	 {Proceedings of Machine Learning Research},
  address = 	 {New York, New York, USA},
  month = 	 {20--22 Jun},
  publisher =    {PMLR},
  url = 	 {https://proceedings.mlr.press/v48/oord16.html}
}

@inproceedings{oord2016conditionalimagegenerationpixelcnn,
 author = {van den Oord, Aaron and Kalchbrenner, Nal and Espeholt, Lasse and kavukcuoglu, koray and Vinyals, Oriol and Graves, Alex},
 booktitle = {Advances in Neural Information Processing Systems},
 editor = {D. Lee and M. Sugiyama and U. Luxburg and I. Guyon and R. Garnett},
 pages = {},
 publisher = {Curran Associates, Inc.},
 title = {Conditional Image Generation with PixelCNN Decoders},
 url = {https://proceedings.neurips.cc/paper_files/paper/2016/file/b1301141feffabac455e1f90a7de2054-Paper.pdf},
 volume = {29},
 year = {2016}
}

@article{yu2022scaling,
  title={Scaling autoregressive models for content-rich text-to-image generation},
  author={Yu, Jiahui and Xu, Yuanzhong and Koh, Jing Yu and Luong, Thang and Baid, Gunjan and Wang, Zirui and Vasudevan, Vijay and Ku, Alexander and Yang, Yinfei and Ayan, Burcu Karagol and others},
  journal={arXiv preprint arXiv:2206.10789},
  volume={2},
  number={3},
  pages={5},
  year={2022}
}

@inproceedings{tian2024visualautoregressivemodelingscalable,
 author = {Tian, Keyu and Jiang, Yi and Yuan, Zehuan and Peng, Bingyue and Wang, Liwei},
 booktitle = {Advances in Neural Information Processing Systems},
 doi = {10.52202/079017-2694},
 editor = {A. Globerson and L. Mackey and D. Belgrave and A. Fan and U. Paquet and J. Tomczak and C. Zhang},
 pages = {84839--84865},
 publisher = {Curran Associates, Inc.},
 title = {Visual Autoregressive Modeling: Scalable Image Generation via Next-Scale Prediction},
 url = {https://proceedings.neurips.cc/paper_files/paper/2024/file/9a24e284b187f662681440ba15c416fb-Paper-Conference.pdf},
 volume = {37},
 year = {2024}
}

@article{qu2025visualautoregressivemodelingimage,
  title={Visual autoregressive modeling for image super-resolution},
  author={Qu, Yunpeng and Yuan, Kun and Hao, Jinhua and Zhao, Kai and Xie, Qizhi and Sun, Ming and Zhou, Chao},
  journal={arXiv preprint arXiv:2501.18993},
  year={2025}
}

@InProceedings{wang2025navigatingimagerestorationvars,
    author    = {Wang, Siyang and Zheng, Naishan and Huang, Jie and Zhao, Feng},
    title     = {Navigating Image Restoration with VAR's Distribution Alignment Prior},
    booktitle = {Proceedings of the IEEE/CVF Conference on Computer Vision and Pattern Recognition (CVPR)},
    month     = {June},
    year      = {2025},
    pages     = {7559-7569}
}

@InProceedings{wei2020componentdivideandconquerrealworldimage,
author="Wei, Pengxu
and Xie, Ziwei
and Lu, Hannan
and Zhan, Zongyuan
and Ye, Qixiang
and Zuo, Wangmeng
and Lin, Liang",
editor="Vedaldi, Andrea
and Bischof, Horst
and Brox, Thomas
and Frahm, Jan-Michael",
title="Component Divide-and-Conquer for Real-World Image Super-Resolution",
booktitle="Computer Vision -- ECCV 2020",
year="2020",
publisher="Springer International Publishing",
address="Cham",
pages="101--117",
isbn="978-3-030-58598-3"
}

@article{loshchilov2019decoupledweightdecayregularization,
  title={Decoupled weight decay regularization},
  author={Loshchilov, Ilya and Hutter, Frank},
  journal={arXiv preprint arXiv:1711.05101},
  year={2017}
}

@InProceedings{rombach2022highresolutionimagesynthesislatent,
    author    = {Rombach, Robin and Blattmann, Andreas and Lorenz, Dominik and Esser, Patrick and Ommer, Bj\"orn},
    title     = {High-Resolution Image Synthesis With Latent Diffusion Models},
    booktitle = {Proceedings of the IEEE/CVF Conference on Computer Vision and Pattern Recognition (CVPR)},
    month     = {June},
    year      = {2022},
    pages     = {10684-10695}
}

@InProceedings{zhang2018unreasonableeffectivenessdeepfeatures,
author = {Zhang, Richard and Isola, Phillip and Efros, Alexei A. and Shechtman, Eli and Wang, Oliver},
title = {The Unreasonable Effectiveness of Deep Features as a Perceptual Metric},
booktitle = {Proceedings of the IEEE Conference on Computer Vision and Pattern Recognition (CVPR)},
month = {June},
year = {2018}
}

@ARTICLE{Ding_2020,
  author={Ding, Keyan and Ma, Kede and Wang, Shiqi and Simoncelli, Eero P.},
  journal={IEEE Transactions on Pattern Analysis and Machine Intelligence}, 
  title={Image Quality Assessment: Unifying Structure and Texture Similarity}, 
  year={2022},
  volume={44},
  number={5},
  pages={2567-2581},
  doi={10.1109/TPAMI.2020.3045810}}

@article{wang2022exploringclipassessinglook, title={Exploring CLIP for Assessing the Look and Feel of Images}, volume={37}, url={https://ojs.aaai.org/index.php/AAAI/article/view/25353}, DOI={10.1609/aaai.v37i2.25353}, abstractNote={Measuring the perception of visual content is a long-standing problem in computer vision. Many mathematical models have been developed to evaluate the look or quality of an image. Despite the effectiveness of such tools in quantifying degradations such as noise and blurriness levels, such quantification is loosely coupled with human language. When it comes to more abstract perception about the feel of visual content, existing methods can only rely on supervised models that are explicitly trained with labeled data collected via laborious user study. In this paper, we go beyond the conventional paradigms by exploring the rich visual language prior encapsulated in Contrastive Language-Image Pre-training (CLIP) models for assessing both the quality perception (look) and abstract perception (feel) of images without explicit task-specific training. In particular, we discuss effective prompt designs and show an effective prompt pairing strategy to harness the prior. We also provide extensive experiments on controlled datasets and Image Quality Assessment (IQA) benchmarks. Our results show that CLIP captures meaningful priors that generalize well to different perceptual assessments.}, number={2}, journal={Proceedings of the AAAI Conference on Artificial Intelligence}, author={Wang, Jianyi and Chan, Kelvin C.K. and Loy, Chen Change}, year={2023}, month={Jun.}, pages={2555-2563} }

@InProceedings{ke2021musiqmultiscaleimagequality,
    author    = {Ke, Junjie and Wang, Qifei and Wang, Yilin and Milanfar, Peyman and Yang, Feng},
    title     = {MUSIQ: Multi-Scale Image Quality Transformer},
    booktitle = {Proceedings of the IEEE/CVF International Conference on Computer Vision (ICCV)},
    month     = {October},
    year      = {2021},
    pages     = {5148-5157}
}

@InProceedings{yang2022maniqamultidimensionattentionnetwork,
    author    = {Yang, Sidi and Wu, Tianhe and Shi, Shuwei and Lao, Shanshan and Gong, Yuan and Cao, Mingdeng and Wang, Jiahao and Yang, Yujiu},
    title     = {MANIQA: Multi-Dimension Attention Network for No-Reference Image Quality Assessment},
    booktitle = {Proceedings of the IEEE/CVF Conference on Computer Vision and Pattern Recognition (CVPR) Workshops},
    month     = {June},
    year      = {2022},
    pages     = {1191-1200}
}

@ARTICLE{1284395,
  author={Zhou Wang and Bovik, A.C. and Sheikh, H.R. and Simoncelli, E.P.},
  journal={IEEE Transactions on Image Processing}, 
  title={Image quality assessment: from error visibility to structural similarity}, 
  year={2004},
  volume={13},
  number={4},
  pages={600-612},
  doi={10.1109/TIP.2003.819861}}

@article{ren2026hallucinationscoremitigatinghallucinations,
  title={Hallucination Score: Towards Mitigating Hallucinations in Generative Image Super-Resolution},
  author={Ren, Weiming and Goyal, Raghav and Hu, Zhiming and Aumentado-Armstrong, Tristan Ty and Mohomed, Iqbal and Levinshtein, Alex},
  journal={arXiv preprint arXiv:2507.14367},
  year={2025}
}

@ARTICLE{mittal2012making,
  author={Mittal, Anish and Soundararajan, Rajiv and Bovik, Alan C.},
  journal={IEEE Signal Processing Letters}, 
  title={Making a “Completely Blind” Image Quality Analyzer}, 
  year={2013},
  volume={20},
  number={3},
  pages={209-212},
  doi={10.1109/LSP.2012.2227726}}

@article{jiao2026flexvarflexiblevisualautoregressive,
  title={Flexvar: Flexible visual autoregressive modeling without residual prediction},
  author={Jiao, Siyu and Zhang, Gengwei and Qian, Yinlong and Huang, Jiancheng and Zhao, Yao and Shi, Humphrey and Ma, Lin and Wei, Yunchao and Jie, Zequn},
  journal={arXiv preprint arXiv:2502.20313},
  year={2025}
}

@article{zhou2025rethinkingtrainingdynamicsscalewise,
  title={Rethinking Training Dynamics in Scale-wise Autoregressive Generation},
  author={Zhou, Gengze and Ge, Chongjian and Tan, Hao and Liu, Feng and Hong, Yicong},
  journal={arXiv preprint arXiv:2512.06421},
  year={2025}
}

@article{tang2024hartefficientvisualgeneration,
  title={Hart: Efficient visual generation with hybrid autoregressive transformer},
  author={Tang, Haotian and Wu, Yecheng and Yang, Shang and Xie, Enze and Chen, Junsong and Chen, Junyu and Zhang, Zhuoyang and Cai, Han and Lu, Yao and Han, Song},
  journal={arXiv preprint arXiv:2410.10812},
  year={2024}
}

@InProceedings{han2025infinityscalingbitwiseautoregressive,
    author    = {Han, Jian and Liu, Jinlai and Jiang, Yi and Yan, Bin and Zhang, Yuqi and Yuan, Zehuan and Peng, Bingyue and Liu, Xiaobing},
    title     = {Infinity: Scaling Bitwise AutoRegressive Modeling for High-Resolution Image Synthesis},
    booktitle = {Proceedings of the IEEE/CVF Conference on Computer Vision and Pattern Recognition (CVPR)},
    month     = {June},
    year      = {2025},
    pages     = {15733-15744}
}

@misc{ho2022classifierfreediffusionguidance,
      title={Classifier-Free Diffusion Guidance}, 
      author={Jonathan Ho and Tim Salimans},
      year={2022},
      eprint={2207.12598},
      archivePrefix={arXiv},
      primaryClass={cs.LG},
      url={https://arxiv.org/abs/2207.12598}, 
}

@InProceedings{qu2024xpsrcrossmodalpriorsdiffusionbased,
author="Qu, Yunpeng
and Yuan, Kun
and Zhao, Kai
and Xie, Qizhi
and Hao, Jinhua
and Sun, Ming
and Zhou, Chao",
editor="Leonardis, Ale{\v{s}}
and Ricci, Elisa
and Roth, Stefan
and Russakovsky, Olga
and Sattler, Torsten
and Varol, G{\"u}l",
title="XPSR: Cross-Modal Priors for Diffusion-Based Image Super-Resolution",
booktitle="Computer Vision -- ECCV 2024",
year="2025",
publisher="Springer Nature Switzerland",
address="Cham",
pages="285--303",
isbn="978-3-031-73247-8"
}

@InProceedings{zhang2023addingconditionalcontroltexttoimage,
    author    = {Zhang, Lvmin and Rao, Anyi and Agrawala, Maneesh},
    title     = {Adding Conditional Control to Text-to-Image Diffusion Models},
    booktitle = {Proceedings of the IEEE/CVF International Conference on Computer Vision (ICCV)},
    month     = {October},
    year      = {2023},
    pages     = {3836-3847}
}

@misc{ye2023ipadaptertextcompatibleimage,
      title={IP-Adapter: Text Compatible Image Prompt Adapter for Text-to-Image Diffusion Models}, 
      author={Hu Ye and Jun Zhang and Sibo Liu and Xiao Han and Wei Yang},
      year={2023},
      eprint={2308.06721},
      archivePrefix={arXiv},
      primaryClass={cs.CV},
      url={https://arxiv.org/abs/2308.06721}, 
}

@InProceedings{wang2018sftgan,
author = {Wang, Xintao and Yu, Ke and Dong, Chao and Loy, Chen Change},
title = {Recovering Realistic Texture in Image Super-Resolution by Deep Spatial Feature Transform},
booktitle = {Proceedings of the IEEE Conference on Computer Vision and Pattern Recognition (CVPR)},
month = {June},
year = {2018}
}

@InProceedings{Cai_2019_ICCV,
author = {Cai, Jianrui and Zeng, Hui and Yong, Hongwei and Cao, Zisheng and Zhang, Lei},
title = {Toward Real-World Single Image Super-Resolution: A New Benchmark and a New Model},
booktitle = {Proceedings of the IEEE/CVF International Conference on Computer Vision (ICCV)},
month = {October},
year = {2019}
}

@InProceedings{Agustsson_2017_CVPR_Workshops,
author = {Agustsson, Eirikur and Timofte, Radu},
title = {NTIRE 2017 Challenge on Single Image Super-Resolution: Dataset and Study},
booktitle = {Proceedings of the IEEE Conference on Computer Vision and Pattern Recognition (CVPR) Workshops},
month = {July},
year = {2017}
}

@inproceedings{hu2026semi,
  title={Semi-supervised latent disentangled diffusion model for textile pattern generation},
  author={Hu, Chenggong and Wang, Yi and Xue, Mengqi and Zhang, Haofei and Song, Jie and Sun, Li},
  booktitle={Proceedings of the AAAI Conference on Artificial Intelligence},
  volume={40},
  number={6},
  pages={4798--4806},
  year={2026}
}

@misc{hu2026gemtalk, title = {Geometry-guided Emotion Modulation for Controllable and Photorealistic Emotional Talking Face Generation}, author = {Chenggong Hu and Shaoyin Ma and Yi Wang and Li Sun and Mingli Song and Jie Song}, year = {2026}, eprint = {2608.00663}, archivePrefix= {arXiv}, primaryClass = {cs.CV}, doi = {10.48550/arXiv.2608.00663}, url = {https://arxiv.org/abs/2608.00663} }

\end{document}